%% file: main.tex
\documentclass[letterpaper,10pt,conference]{ieeeconf}
\IEEEoverridecommandlockouts
\usepackage[T1]{fontenc}
\usepackage[utf8]{inputenc}
\usepackage{graphicx}
\usepackage{amsmath,amssymb}
\usepackage{booktabs,array}
\usepackage{placeins}
\usepackage{cite}
\usepackage{url}
\usepackage[hidelinks]{hyperref}
\graphicspath{{figures/}}

\hypersetup{
  pdftitle={Confidence Filtering in Video-Based Gait Assessment: Measurement Error versus Selection Shift},
  pdfauthor={David Regev and Eran Beeri Bamani},
  pdfsubject={Preprint submitted to IEEE ICORR 2027}
}

\title{\LARGE\bf
Confidence Filtering in Video-Based Gait Assessment: Measurement Error versus Selection Shift}

\author{David Regev$^{1}$ and Eran Beeri Bamani$^{1}$%
\thanks{$^{1}$Department of Information Science and Applied Artificial Intelligence,
Bar-Ilan University, Ramat Gan, Israel.}%
\thanks{Submitted to the 2027 IEEE International Conference on Rehabilitation Robotics (ICORR). This work has been submitted to the IEEE for possible publication. Copyright may be transferred without notice, after which this version may no longer be accessible.}%
}

\begin{document}
\bstctlcite{IEEEexample:BSTcontrol}
\maketitle
\thispagestyle{empty}
\pagestyle{empty}

\begin{abstract}
Video-based gait assessment often keeps only high-confidence pose observations, which can lower measurement error but change the movement assessed. We compared confidence selection with budget-matched random selection in a Parkinson-related cohort (n = 96) and an NF2-related replication cohort (n = 18). In the Parkinson-related cohort, confidence selection produced a small but consistent shift in the observed movement summary (+0.05 SD, 95\% CI 0.03 to 0.07), while held-out Mini-BESTest error stayed flat ($-0.04$ points, $-0.13$ to 0.05); a policy preserving motion-group proportions reduced the shift, while NF2 showed no shift and its predictor did not beat a constant baseline. Because this shift mixes measurement error with selection shift, we decompose and simulate both on the cohort's window structure. Filtering lowers local error when error depends on confidence, but biases the endpoint once confidence tracks movement more than error. In the Parkinson-related data, confidence was only weakly tied to movement ($\rho$ = 0.14). There, filtering is harmless if well-tracked windows are unbiased. If instead systematic error only redistributes within participants, filtering can add up to 0.58$^\circ$. Unchanged clinical error from such small-cohort predictors therefore cannot show that quality filtering preserves the assessed movement.
\end{abstract}

\input{introduction}
\input{methods}
\input{evaluation}

\input{Discussion}
\input{conclusions}

\IEEEtriggeratref{0}
\bibliographystyle{IEEEtran}
\bibliography{references}

\clearpage
\onecolumn
\section*{Appendix: Supplementary simulation results}
\setcounter{figure}{0}
\renewcommand{\thefigure}{S\arabic{figure}}
\renewcommand{\theHfigure}{S\arabic{figure}}
\begin{figure}[h]
\centering
\includegraphics[width=\textwidth]{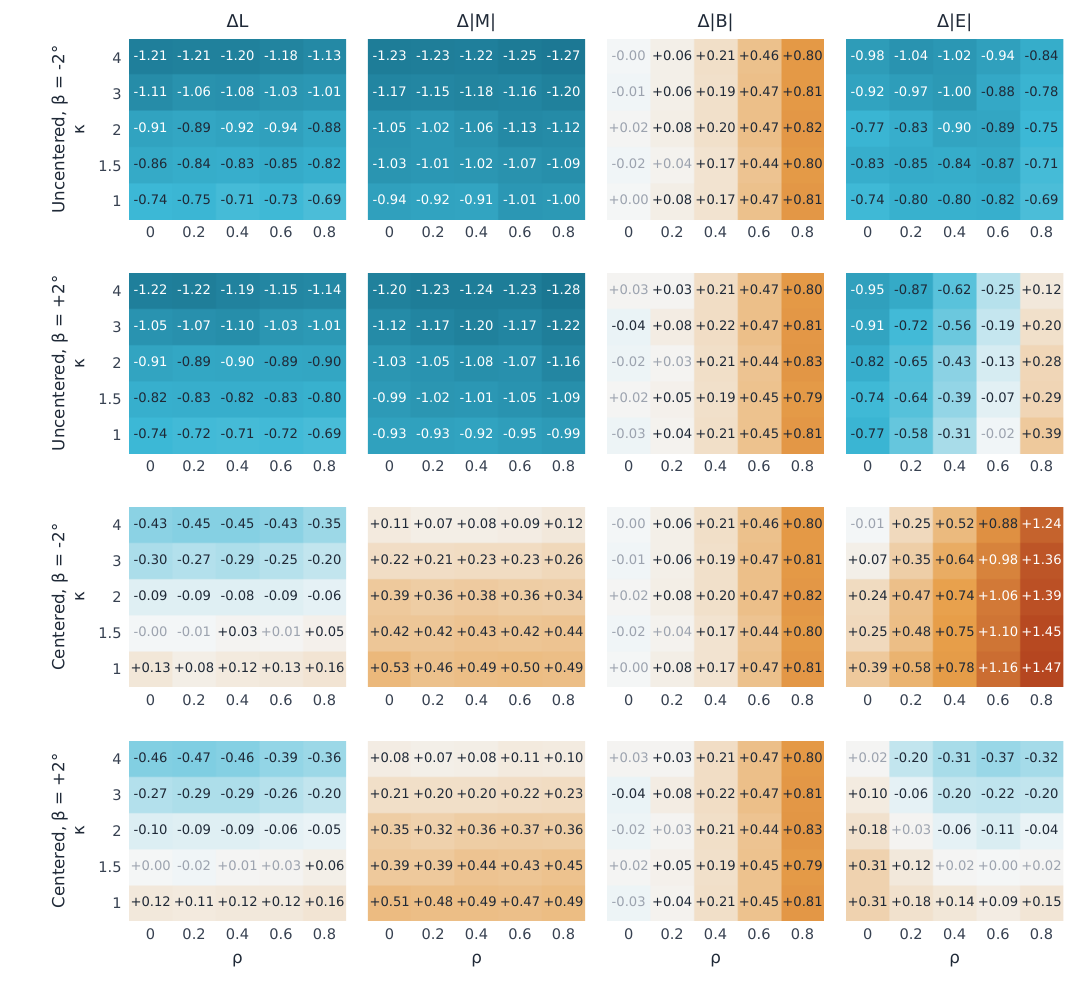}
\caption{Confidence-linked systematic error at $\sigma_0=2^\circ$: change from confidence relative to random selection at 50\% nominal coverage (degrees; grey: within two standard errors of zero) in local error $L_i$, $|M_i|$, $|B_i|$ and $|E_i|$, by $\rho$ and $\kappa$. Uncentered: error mean $-\beta(1-c_{iu})$, so the best-tracked windows are unbiased. Centered: error mean $-\beta(\bar c_i-c_{iu})$, so bias only redistributes error within participant.}
\label{fig:s1}
\end{figure}
\begin{figure}[h]
\centering
\includegraphics[width=\textwidth]{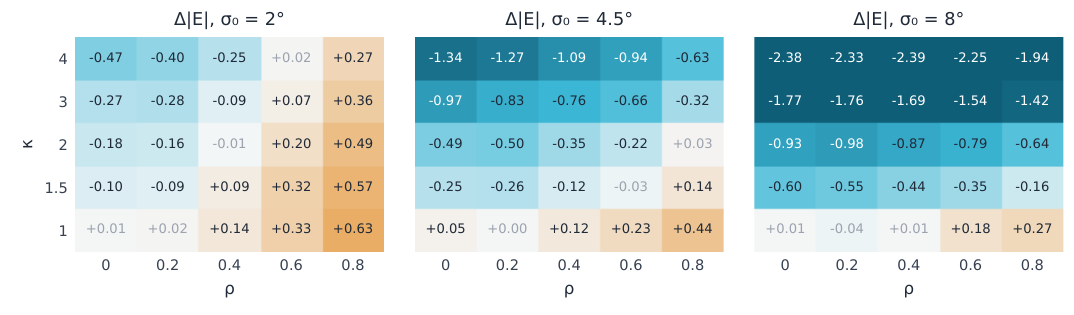}
\caption{Zero-mean error ($\beta=0$): change in $|E_i|$ from confidence relative to random selection at 50\% nominal coverage across baseline error $\sigma_0$ (degrees; grey: within two standard errors of zero). $\sigma_0=8^\circ$ exceeds the observed within-participant SD of window knee excursion (4.35$^\circ$).}
\label{fig:s2}
\end{figure}
\end{document}

%% file: introduction.tex
\section{Introduction}

\begin{figure*}[t]
  \centering
  \includegraphics[width=\textwidth]{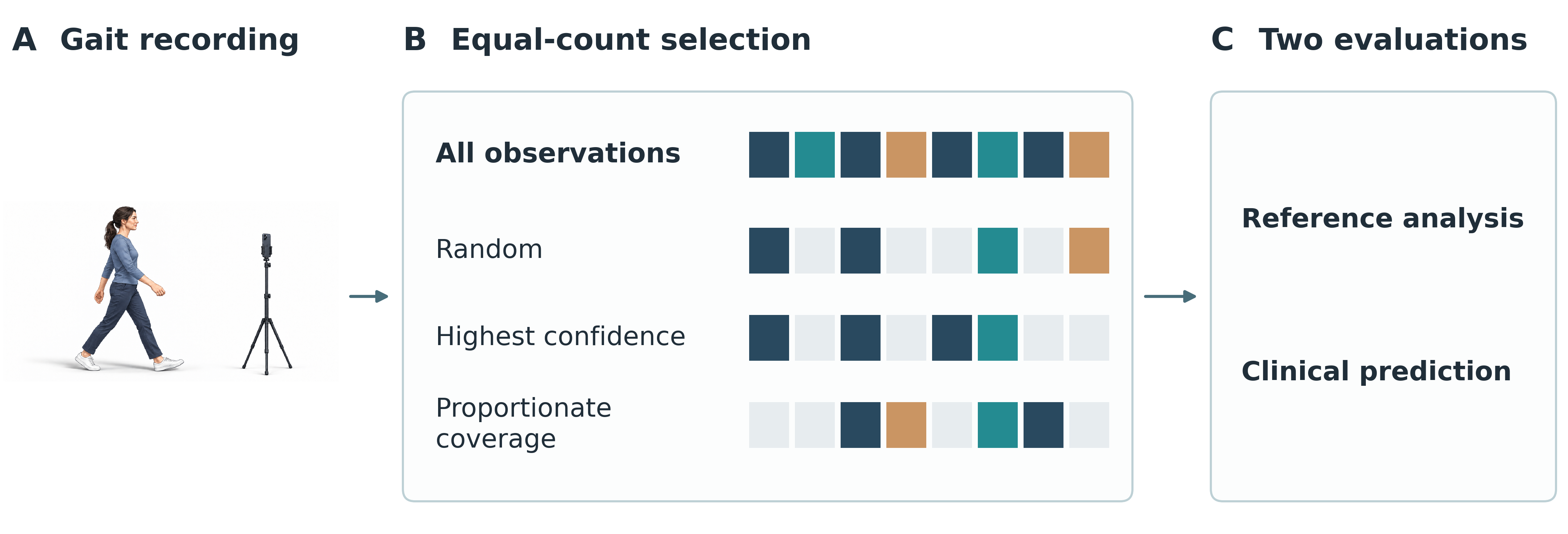}
  \caption{Observation selection at a matched budget. (A) All policies share one eligible recording. (B) Illustrative masks retain four of eight units; colors indicate motion groups and gray marks omission. Units are pose windows in the clinical cohorts (complete strides when a paired reference is available); motion centrality is omitted for clarity. (C) A paired reference separates measurement error from selection shift; without one, the clinical cohorts yield the observed shift and held-out Mini-BESTest prediction error.}
  \label{fig:overview}
  \vspace{-0.4cm}
\end{figure*}

Video-based movement analysis enables quantitative gait assessment outside the laboratory. Single-camera models estimate gait measures and clinical indices in cerebral palsy~\cite{Kidzinski2020,Zhao2024,Bamani2026DiffusionGait}, and multiview smartphone systems recover kinematics against laboratory references~\cite{Uhlrich2023}. Pose-based gait analysis has been validated in the image plane~\cite{Stenum2021} and across camera views, clinical populations, and sensitivity to change~\cite{Stenum2024}, and learned gait representations have been related to biomechanical descriptors~\cite{BeeriBioRob2026}. For rehabilitation, however, what matters is the validity of the person-level summary, not only the accuracy of each pose estimate: home monitoring and robot-assisted therapy increasingly track progress from such summaries, and a summary that shifts with data quality can mimic or mask change.

Which observations should form that summary? Low-confidence detections and occlusions degrade gait estimates, and markerless capture shows higher inter-trial variability than marker-based capture~\cite{Horsak2024}. Yet difficult-to-track movement can be part of the behavior being assessed: atypical appearance challenges detectors~\cite{Zhou2025}, and movement uncertainty itself has been used to detect gait abnormality~\cite{Chakraborty2020}. Discarding low-confidence observations can therefore reduce measurement error while shifting the sample of movement summarized. Calibrated uncertainty~\cite{Donahue2026} addresses whether a confidence score tracks reconstruction error, but not whether the retained observations represent the full record. Selective prediction shows a related risk: aggregate gains after abstention can hide worse performance for subgroups~\cite{Jones2021,Shah2022}. Here selection operates within a person; every participant remains in the analysis.

Isolating this effect requires two controls. First, the number of observations must be matched, because the number and choice of gait cycles alone alter measurement error~\cite{Fox2023}; a selected subset must be compared with a random subset of the same size, not with the full record. Second, an independent reference is needed to separate measurement error from genuine movement variation.

We ask whether selecting high-quality observations reduces local measurement error while shifting the person-level endpoint away from an independent full-record reference (Fig.~\ref{fig:overview}). The contributions are:

\begin{itemize}
\item
  In a Parkinson-related cohort, confidence selection shifted the observed movement summary more than budget-matched random selection while held-out Mini-BESTest error stayed flat, and a candidate proportionate-coverage policy reduced that shift; an NF2-related schwannomatosis cohort showed no selection shift.
\item
  A within-participant decomposition of full-record endpoint error into measurement error and selection shift, an accounting identity used as an evaluation tool rather than a new estimator.
\item
  A simulation with known components on the clinical window structure that explains when confidence filtering helps and when it harms the endpoint, and places the Parkinson-related pipeline on that map.
\end{itemize}

%% file: methods.tex
\section{Methods}

\subsection{Problem formulation}

For participant $i$, let $\mathcal{U}_i$ be the eligible movement units for a task. With a paired reference, a unit would be a complete stride with synchronized markerless and reference measurements; in the clinical cohorts it is a non-overlapping pose window. Eligibility requires a valid time span, participant identity, and the data needed to compute the unit endpoint. It is fixed before selection and shared by all policies, and low confidence alone is not an exclusion criterion. Units are grouped into strata, one per task, walking direction, or acquisition setting. Every policy retains the same number of units from each stratum, and strata are combined with fixed weights $\omega_{is}=N_{is}/\sum_s N_{is}$, so all policies share the same stratum mix and the target remains the full eligible-unit mean. Equations~\eqref{eq:means}--\eqref{eq:local} are written for a single stratum and applied within each stratum.

\subsection{Separating measurement error from selection shift}

\label{sec:decomposition}

Let $z_{iu}$ be the reference endpoint of unit $u$ and $\hat z_{iu}$ its markerless estimate. For a retained set $\mathcal{A}_i\subseteq\mathcal{U}_i$,

\begin{equation}
\mu^Z_{\mathcal{A}_i}=\frac{1}{|\mathcal{A}_i|}\sum_{u\in\mathcal{A}_i}z_{iu},\qquad \mu^X_{\mathcal{A}_i}=\frac{1}{|\mathcal{A}_i|}\sum_{u\in\mathcal{A}_i}\hat z_{iu}.
\label{eq:means}
\end{equation}

\begin{figure*}[t]
  \centering
  \includegraphics[width=\textwidth]{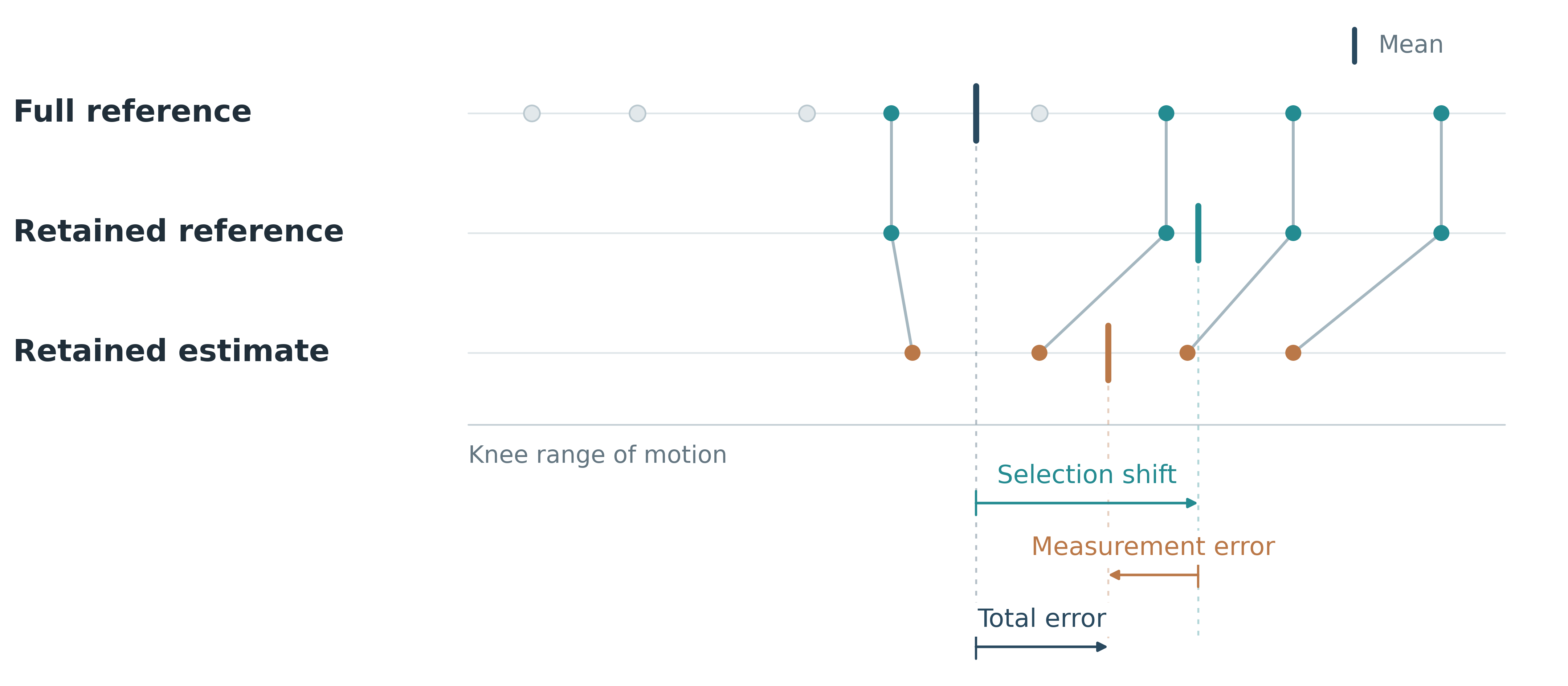}
  \caption{Separating selection shift from measurement error. Dots show stride-level knee ranges of motion; vertical bars mark means. Teal and gray denote retained and omitted reference strides; orange denotes markerless estimates. Links match strides across rows. Arrows indicate signed differences between means. Values and selections are illustrative.}
  \label{fig:decomposition}
  \vspace{-0.2cm}
\end{figure*}

The signed error relative to the full reference record decomposes as

\begin{equation}
E_i=\mu^X_{\mathcal{A}_i}-\mu^Z_{\mathcal{U}_i}=\underbrace{\mu^X_{\mathcal{A}_i}-\mu^Z_{\mathcal{A}_i}}_{M_i}+\underbrace{\mu^Z_{\mathcal{A}_i}-\mu^Z_{\mathcal{U}_i}}_{B_i},
\label{eq:decomp}
\end{equation}

where $M_i$ is aggregate measurement error on retained units and $B_i$ is the selection shift of the reference endpoint (Fig.~\ref{fig:decomposition}), an observed finite-record difference that on its own does not establish population bias or clinical harm. Positive and negative errors on individual strides can cancel in this mean, so a small $|M_i|$ need not imply accurate strides. Local measurement quality is therefore assessed separately by

\begin{equation}
L_i=\frac{1}{|\mathcal{A}_i|}\sum_{u\in\mathcal{A}_i}|\hat z_{iu}-z_{iu}|.
\label{eq:local}
\end{equation}

The endpoint is mean knee excursion (degrees): the within-unit range of the image-plane knee angle, averaged over sides and computed for each unit before selection; the observed shift below keeps sides separate. With a paired laboratory reference, the same decomposition would use 3D knee-flexion range of motion from both systems, because 2D angles depend on camera viewpoint.

\subsection{Matched budgets and selection policies}

At coverage $q$, each policy retains $n_{is}(q)=\lfloor qN_{is}\rfloor$ units per stratum, with $q=0.5$ compared against the full record ($q=1$). Participants need at least four eligible units and one retained unit per stratum. Clinical strata hold few windows (2 to 3 in the Parkinson-related cohort), so lower coverages would empty strata and are not evaluated; flooring makes realized coverage at $q=0.5$ 38\% (Parkinson-related) and 46\% (NF2). The budget is a unit count; for fixed-length clinical windows this also fixes the frame count, while retained stride duration may vary and is reported. All analyses are offline.

\emph{Random selection} samples units without replacement within strata; losses, not predictions, are averaged over $R=100$ draws per participant. \emph{Confidence selection} ranks units by the 10th percentile of keypoint confidence as stored by each pipeline (for NF2, after the short-gap filling of~\cite{Bamani2026Bridge}, which assigns $c=0.2$ to filled keypoints) over hip, knee, and ankle detections, within each participant and pose pipeline, with ties broken by a fixed identifier order. \emph{Motion centrality} retains, within each stratum, the units with the smallest Euclidean distance to the median descriptor vector of that stratum's units, in standardized feature space. \emph{Proportionate coverage} assigns each unit to one of $K$ motion clusters by k-means~\cite{Lloyd1982} and applies confidence ranking within each cluster. With $N_{isk}$ denoting the number of the $N_{is}$ units assigned to cluster $k\in\{1,\dots,K\}$, each cluster receives $n_{is}(q)\cdot N_{isk}/N_{is}$ of the stratum budget, rounded to integers by largest remainder so the quotas sum to $n_{is}(q)$; rare clusters thus keep their observed share rather than equal weight. Selectors use eight per-window image-plane descriptors from the shoulder, hip, knee, and ankle keypoints, without interpolation (a keypoint is missing when its confidence is zero or non-finite, and each descriptor requires at least eight valid frames): the median and 5th-to-95th-percentile range (excursion) of the left and right knee angle; the median and interquartile range of inter-ankle distance normalized by median hip width; and the median and interquartile range of trunk inclination from the image vertical. Descriptors are standardized with training-participant statistics after median imputation, and $k$-means centers with $K\in\{2,3,4\}$ chosen by silhouette score are fitted on training participants only. Both clinical archives were recorded with a fixed monocular camera lateral to the walkway (approximately sagittal view)~\cite{Bamani2026Bridge}. Parkinson-related poses were provided as OpenPose COCO-17 arrays, and NF2 poses were extracted with MediaPipe Pose and mapped to COCO-17~\cite{Bamani2026Bridge}; confidence scales are estimator-specific, so ranking is done within each pipeline. Reference trajectories and clinical labels of evaluation participants are never selector inputs.

\begin{table}[ht]
\caption{Data sources and evaluation roles.}
\label{tab:data}
\centering
\small
\setlength{\tabcolsep}{3pt}
\renewcommand{\arraystretch}{1.12}
\begin{tabular}{@{}>{\raggedright\arraybackslash}p{0.19\columnwidth}
>{\raggedright\arraybackslash}p{0.44\columnwidth}
>{\raggedright\arraybackslash}p{0.31\columnwidth}@{}}
\toprule
Source & Analyzed inventory & Evaluation role \\
\midrule
Parkinson-related~\cite{Bamani2026Bridge,MIT77PD} & 96 of 145 participants; 763 windows, $124\times17\times3$; 3 dual-task strata & Observed shift; primary clinical prediction \\
Simulation & Parkinson-related window structure; synthetic knee-excursion truth & $M$, $B$, $E$, $L$ known \\
NF2~\cite{Bamani2026Bridge,Stec2026} & 18 of 19 participants; 175 windows, $64\times17\times3$; outbound/return strata & Clinical replication \\
\bottomrule
\end{tabular}
\vspace{-0.4cm}
\end{table}

%% file: evaluation.tex
\section{Evaluation and Results}

\subsection{Data sources and provenance}

Table~\ref{tab:data} summarizes the data. \textbf{Clinical cohorts.} The Parkinson-related archive (AMPARO network~\cite{Bamani2026Bridge,MIT77PD}; 107 people with idiopathic Parkinson's disease and 38 age-matched controls; 96 analyzed) contains participant-level Mini-BESTest scores (range 10 to 28) and 763 image-plane pose windows, one randomly chosen (seeded) clip per walking recording, $X\in\mathbb{R}^{124\times17\times3}$ (124 frames, 17 keypoints, and x, y, confidence per keypoint). The Mini-BESTest assesses several components of dynamic balance~\cite{Franchignoni2010}, so a gait-based prediction estimates a broader score. Participants performed three standardized dual-task walking conditions~\cite{Bamani2026Bridge} (simple walking, walking while counting down, and walking with verbal fluency), which are treated as strata; NF2 participants performed a 3-m walk-and-return, with outbound and return bouts as strata~\cite{Bamani2026Bridge}. Each participant contributes one visit with its linked Mini-BESTest assessment. The NF2-related schwannomatosis archive~\cite{Bamani2026Bridge,Stec2026} provides a prespecified replication with windows $X\in\mathbb{R}^{64\times17\times3}$, scores from 6 to 27 averaged over two raters, and 175 non-augmented windows from 18 participants; augmented windows are excluded. All units are linked to participant, visit, source recording, task, and frame interval.

\subsection{Held-out clinical prediction}

An RBF-kernel support vector regressor ($C=1$, $\gamma=0.1$, $\epsilon\in\{0.1,0.5\}$), selected over ridge regression by inner validation in every outer fold, predicts the Mini-BESTest total from the eight selector descriptors plus the interquartile range of each knee angle, averaged with the fixed stratum weights; confidence drives selection but is not a predictor. The PD cohort uses five participant-grouped outer folds and NF2 uses nested leave-one-participant-out evaluation. All windows of a participant stay in one fold; scaling, regularization, and clusters are fitted on training data, and inner validation uses participant-level loss with one feature vector per participant. In the primary analysis the predictor is trained on full eligible records and held fixed while each selector is applied to held-out records. A sensitivity analysis refits the predictor under each policy with the same outer splits.

\begin{table}[t]
\centering
\caption{Mini-BESTest mean absolute error (MAE), in points, under observation selection at matched nominal coverage (frozen predictor). $\Delta$ rows are policy minus random with 95\% participant-bootstrap intervals.}
\label{tab:clinical_selection}
\small
\setlength{\tabcolsep}{3pt}
\renewcommand{\arraystretch}{1.12}
\begin{tabular*}{\columnwidth}{@{\extracolsep{\fill}}lcc@{}}
\toprule
\textbf{Method}
& \shortstack{\textbf{Parkinson-related}\\MAE $\downarrow$}
& \shortstack{\textbf{NF2}\\MAE $\downarrow$} \\
\midrule
Training mean & 3.02 & 5.98 \\
Training median & 3.04 & 4.97 \\
Full record (100\%) & 2.83 & 5.32 \\
\midrule
\multicolumn{3}{@{}l}{\textit{50\% nominal coverage}} \\
Random & 2.87 & 5.29 \\
Confidence & 2.84 & 5.29 \\
Motion centrality & 2.86 & 5.11 \\
Proportionate & 2.88 & 5.29 \\
\midrule
\multicolumn{3}{@{}l}{\textit{$\Delta$ vs.\ Random, 50\% coverage [95\% CI]}} \\
Confidence & $-0.04$ {\footnotesize[$-0.13$, $0.05$]} & $-0.00$ {\footnotesize[$-0.18$, $0.19$]} \\
Centrality & $-0.01$ {\footnotesize[$-0.07$, $0.05$]} & $-0.18$ {\footnotesize[$-0.29$, $-0.07$]} \\
Proportionate & $+0.00$ {\footnotesize[$-0.07$, $0.08$]} & $-0.00$ {\footnotesize[$-0.18$, $0.19$]} \\
\bottomrule
\end{tabular*}
\vspace{-0.4cm}
\end{table}

\subsection{Outcomes and inference}

In the simulation, where the reference is known, the primary contrast is the participant-level difference in absolute full-record endpoint error between confidence and random selection at 50\% coverage,

\begin{equation}
\Delta^E_i=|E_i(\mathcal{A}^{\mathrm{conf}}_i)|-\frac{1}{R}\sum_{r=1}^{R}|E_i(\mathcal{A}^{\mathrm{rand}}_{ir})|,
\label{eq:delta}
\end{equation}

with matching contrasts for $L_i$ and $|B_i|$. In the clinical cohorts, which have no reference and therefore no $E_i$, the primary outcome is the same confidence-versus-random comparison in held-out Mini-BESTest MAE. Means and 95\% intervals use 2,000 participant-level bootstrap resamples with fixed seeds; folds, windows, and draws are not treated as independent participants. For cross-validated predictions, intervals condition on the fitted folds. Without a reference, the clinical cohorts cannot separate $M_i$ from $B_i$; we report the observed shift $S_i=\mu^X_{\mathcal{A}_i}-\mu^X_{\mathcal{U}_i}$ of the eight descriptors (mean absolute value, training-SD units), which combines both: lower knee excursion in high-confidence windows could reflect better tracking of a less-flexed knee rather than different movement. Other coverage levels and policies are reported descriptively, without multiplicity adjustment. \emph{Simulation.} To instantiate the decomposition where $M_i$ and $B_i$ are known, we ran an illustrative simulation (not prespecified), treating the mean 2D knee excursion of each Parkinson-related window as synthetic truth $z_{iu}$, keeping the real participant and stratum structure. Confidence was generated within participant as $c_{iu}=\rho(-\tilde z_{iu})+\sqrt{1-\rho^2}\,\xi_{iu}$, with $\tilde z$ the within-participant standardized truth and $\xi\sim\mathcal{N}(0,1)$, so $\rho>0$ gives straighter knees higher confidence; estimates were $\hat z_{iu}=z_{iu}+\varepsilon_{iu}$ with $\varepsilon_{iu}\sim\mathcal{N}(-\beta(1-c_{iu}),\sigma_0^2e^{-2\lambda c_{iu}})$, or with mean $-\beta(\bar c_i-c_{iu})$ so that bias only redistributes error within participant, and $\lambda=\ln\kappa/2$, so $\kappa$ is the error ratio between low- and high-confidence windows and $\beta$ a systematic error that grows as confidence falls ($\beta>0$: poorly tracked windows under-read excursion, as under occlusion; $\beta<0$: they over-read it, as when jitter inflates the range). We swept $\rho\in[0,0.8]$, $\kappa\in[1,4]$, $\beta\in\{-2,0,2\}^\circ$, and $\sigma_0\in\{2,4.5,8\}^\circ$, bracketing the 4.2$^\circ$ knee-flexion RMSE of two-camera OpenCap~\cite{Uhlrich2023} (single-camera 2D tracking is expected to err more), with 30 replicates per cell (standard error at most 0.09$^\circ$). To place the cohort on the $\rho$ axis, we computed the within-participant correlation between confidence and observed $-z$; random error attenuates it toward zero, and systematic error biases it in the direction of $\beta$.

\subsection{Results}

\emph{Cohorts and eligibility.} The analysis set comprised 96 of 145 Parkinson-related participants, each with 6 to 9 windows across the three conditions, and 18 of 19 NF2 participants with 4 to 21 windows; extreme NF2 windows that prevented model convergence were removed, leaving 175 of 422 base windows; removed windows did not have lower confidence than retained ones (median 10th-percentile keypoint confidence 0.585 vs 0.572; Mann--Whitney $p=0.64$). All analyzed participants had at least one retained unit per stratum at 50\% coverage.

\begin{figure*}[!t]
\centering
\makebox[0.5\textwidth][c]{\small\textbf{(a)}}%
\makebox[0.5\textwidth][c]{\small\textbf{(b)}}\par
\vspace{1mm}
\includegraphics[width=\textwidth]{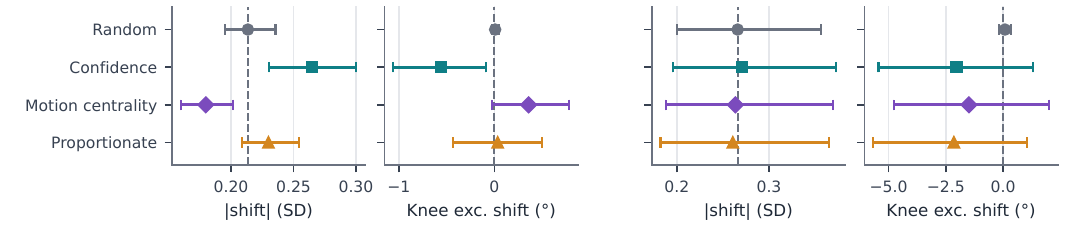}
\caption{Observed selection shift at 50\% nominal coverage: (a) Parkinson-related ($n=96$); (b) NF2 ($n=18$). Left: mean absolute shift of the eight-descriptor movement summary from the full record (training-SD units). Right: signed shift of mean knee excursion (degrees). Bars: 95\% participant-bootstrap intervals; Random is averaged over 100 draws.}
\label{fig:shift}
\vspace{-0.4cm}
\end{figure*}

\emph{Observed selection shift.} Policies retained largely different windows (Jaccard 0.28 to 0.34 against random; 0.76 and 0.71 between confidence and proportionate). In the Parkinson-related cohort, confidence selection shifted the movement summary more than random selection (0.265 vs.~0.214 SD; +0.051, 95\% CI 0.033 to 0.073; 71\% of participants), toward lower knee excursion ($-0.56^\circ$, $-1.05$ to $-0.11$) (Fig.~\ref{fig:shift}). Proportionate coverage reduced this excess ($-0.035$ SD vs.~confidence, $-0.065$ to $-0.013$) leaving +0.016 SD relative to random ($-0.002$ to 0.032); motion centrality shifted it least, as its definition implies. In NF2, no policy differed from random.

\emph{Clinical prediction.} Despite this shift, held-out Mini-BESTest MAE barely moved (Table~\ref{tab:clinical_selection}). In the Parkinson-related cohort, the full-record predictor (2.83 points) improved on the training-median baseline (3.04) by 0.21 points. At 50\% coverage, the four policies ranged from 2.84 to 2.88 points; confidence minus random was $-0.04$ points (95\% CI $-0.13$ to 0.05), and no policy differed from random. Refitting the predictor under each policy left this contrast unchanged ($-0.02$, $-0.09$ to 0.05). In NF2, the full-record predictor (5.32) did not beat the training median (4.97), so NF2 contrasts (Table~\ref{tab:clinical_selection}) indicate direction only: motion centrality was the only policy below random ($-0.18$, $-0.29$ to $-0.07$), and refitting made confidence selection worse than random (+0.32, 0.16 to 0.49). All differences were an order of magnitude below the Mini-BESTest minimal clinically important difference of about 4 points~\cite{Godi2013}. Without a predictor, the Spearman correlation between each descriptor and the Mini-BESTest was weak on full records (Parkinson-related, $|r_s|\le0.37$) and did not differ between any policy and random selection in either cohort.

\emph{Simulation.} Confidence selection reduced local measurement error $L_i$ whenever low-confidence windows were noisier ($\kappa>1$), but increased the selection shift $|B_i|$ as $\rho$ grew, up to 0.8$^\circ$ at $\rho=0.8$ regardless of $\kappa$, $\beta$ or $\sigma_0$. Its net effect on endpoint error $|E_i|$ therefore depended on which of the two dominated (Fig.~\ref{fig:simulation}c). Within participants, observed window knee excursion varied by 4.35$^\circ$ (SD), which caps window-level error, so we report $\sigma_0=2^\circ$ and 4.5$^\circ$ (8$^\circ$ in Fig.~\ref{fig:s2}). At $\sigma_0=2^\circ$, with zero-mean error, confidence selection increased $|E_i|$ once $\rho\ge0.4$ and error depended weakly on confidence (+0.14$^\circ$ at $\kappa=1$; +0.49$^\circ$ at $\rho=0.8$, $\kappa=2$) and reduced it at $\rho\le0.2$ ($-0.16$ to $-0.47^\circ$ for $\kappa\ge2$); at $\sigma_0=4.5^\circ$ harm was confined to $\kappa\le1.5$ (Fig.~\ref{fig:simulation}a,b). The verdict under systematic error depended on which windows it spares (Fig.~\ref{fig:s1}): if the best-tracked windows are unbiased, confidence selection removed most of it, partly through positive $M_i$ cancelling negative $B_i$ when low-confidence windows over-read; if bias only redistributes error around each participant's mean, confidence selection changed $|E_i|$ by $-0.20$ to +0.58$^\circ$ relative to random at $\rho=0.2$ and $\sigma_0=2^\circ$, increasing it for $\kappa\le2$ ($-1.12$ to +0.40$^\circ$ at $4.5^\circ$). The observed Parkinson-related correlation was $\rho=0.14$ (95\% CI 0.05 to 0.22; NF2 $-0.06$); this raw estimate needs no assumption about $\sigma_0$, and correcting it for attenuation at $\sigma_0=2^\circ$ gives 0.15 (0.06 to 0.25). Up to $\rho\approx0.25$, zero-mean cells changed $|E_i|$ by at most about +0.05$^\circ$; near the upper bound on $\sigma_0$, true within-participant variation and hence $\rho$ would vanish. The observed $-0.56^\circ$ excursion shift fits either retained straighter knees ($\rho>0$) or removed jitter-inflated windows ($\beta<0$); only a paired reference can separate them.

%% file: Discussion.tex
\begin{figure*}[t]
\centering
\includegraphics[width=\textwidth]{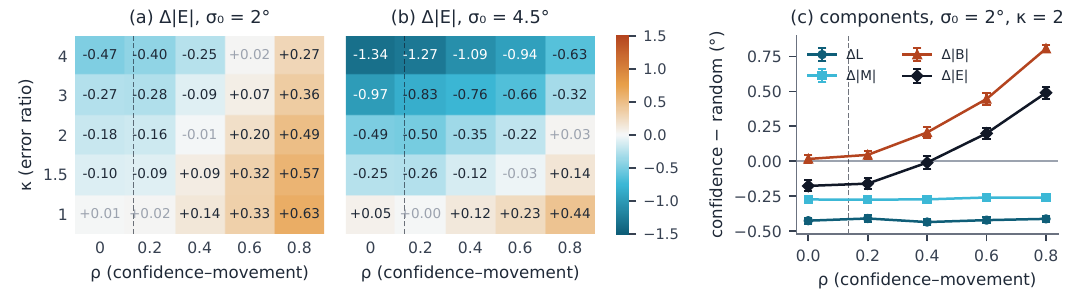}
\caption{Simulation with known components, zero-mean error: change from confidence relative to random selection at 50\% nominal coverage (degrees; orange: confidence selection worse; grey: within two standard errors of zero). (a, b) Absolute full-record endpoint error $|E_i|$ by correlation $\rho$ between confidence and straighter knees and error ratio $\kappa$, at baseline error $\sigma_0=2^\circ$ and $4.5^\circ$. (c) Components at $\sigma_0=2^\circ$, $\kappa=2$ ($\pm2$ SE). Dashed line: observed Parkinson-related $\rho$. 96 synthetic participants, 30 replicates.}
\label{fig:simulation}
\vspace{-0.4cm}
\end{figure*}

\section{Discussion}

In the Parkinson-related cohort, keeping the highest-confidence half of each participant's windows shifted the observed movement summary more than random selection did, toward lower knee excursion, while held-out Mini-BESTest error stayed within 0.04 points of random, more than an order of magnitude below the clinically important difference; proportionate coverage, a candidate rather than a fix, removed about two-thirds of the excess shift. Because the shift is measured on the same 2D estimates, it mixes measurement error with selection shift: it fits retained straighter knees as well as removed jitter-inflated windows.

The simulation explains when such a shift matters. Confidence filtering lowers local error whenever error depends on confidence, but biases the endpoint once confidence tracks movement more than error: with zero-mean error, the net effect crossed zero near $\rho=0.4$ (Fig.~\ref{fig:simulation}c). The Parkinson-related pipeline sits at weak coupling ($\rho\approx0.15$), where zero-mean and unbiased-best-window error predict no harm. That verdict is conditional on the form of systematic error: if well-tracked windows are unbiased, filtering removes bias; if bias only redistributes error around each participant's mean, filtering adds up to 0.58$^\circ$ ($\rho=0.2$, $\kappa=1$, $\beta=-2^\circ$, $\sigma_0=2^\circ$). We consider the first form more plausible, because a confidence score is built to flag inaccurate keypoints, but only a paired reference can show which holds.

A predictor of the kind small clinical cohorts support did not detect the shift: inner validation chose the most regularized model offered, which stays close to the training median. The predictor-free test agreed, however: descriptor--Mini-BESTest correlations did not change under any policy. With two or three windows per condition, a 50\% budget also keeps one, leaving the policies little room to differ.

The bias at stake is small, below 1$^\circ$ here, but directional, so it does not average out across participants and matters most when change is tracked between sessions of differing tracking quality. Studies that filter pose data by quality should compare the filtered subset with a same-size random subset, not the full record, because fewer observations alone change error~\cite{Fox2023}, and should report what the filter kept, because unchanged clinical error does not show that the summary describes the same movement.

\emph{Limitations.} No laboratory reference was available for the clinical cohorts, so measurement error and selection shift were measured only in simulation. The clinical cohorts use 2D image-plane data, where projection error adds to measurement error. The simulation assumes Gaussian error, a single knee-excursion endpoint, and linear confidence--movement coupling, and uses observed 2D excursion as truth, so its map is qualitative; the real values of $\kappa$ and $\beta$ are unknown. The NF2 cohort is small and its predictor did not beat a constant baseline. Widening the regularization grid ($C\ge0.01$, ridge $\alpha\le10^4$) did not improve the near-constant predictors. Confidence was used only as a within-pipeline ranking signal; other estimators or calibrated uncertainties~\cite{Donahue2026} may behave differently.

%% file: conclusions.tex
\section{Conclusion}

In a Parkinson-related cohort, confidence selection shifted the observed gait summary while held-out Mini-BESTest error stayed flat, so clinical error from such predictors is a weak test of whether filtering preserves the assessed movement. A simulation with known components shows that filtering harms the endpoint only when confidence tracks movement more than error; this pipeline sits at weak coupling, where filtering is harmless if well-tracked windows are unbiased. Paired markerless and laboratory validation, for example with the OpenCap release~\cite{Uhlrich2023}, is the next step toward measuring both components directly.

\emph{Data and code availability:} Parkinson-related pose data are publicly available from the AMPARO dataset repository~\cite{MIT77PD}; clinical scores are available from its authors on request. NF2-related schwannomatosis data are not publicly available because of participant privacy. Analysis code will be made publicly available upon acceptance.